\documentclass{article}
\usepackage{makecell}
\usepackage{floatrow}
\usepackage{float}
\usepackage{PRIMEarxiv}
\usepackage{graphicx}
\usepackage[utf8]{inputenc} 
\usepackage[T1]{fontenc}    
\usepackage{hyperref}       
\usepackage{url}            
\usepackage{booktabs}       
\usepackage{amsfonts}       
\usepackage{nicefrac}       
\usepackage{microtype}      
\usepackage{lipsum}
\usepackage{fancyhdr}       
\usepackage{graphicx}       
\graphicspath{{media/}}     
\usepackage{amsmath} 
\usepackage{threeparttable}
\usepackage{array}
\usepackage{footmisc,hyperref}
\usepackage{longtable}
\usepackage{subcaption}
\usepackage{multirow}
\usepackage[table]{xcolor}
\usepackage{multicol}
\usepackage{color}
\usepackage{hyperref}
\usepackage{appendix}
\usepackage{adjustbox}
\usepackage{algorithm}
\usepackage{algpseudocode}
\newtheorem{definition}{Definition}
\newtheorem{theorem}{Theorem}[section]

\newtheorem{remark}[theorem]{Remark}

\hypersetup{
colorlinks=true,
linkcolor=red,
filecolor=cyan,
urlcolor=cyan,
citecolor=cyan,
}
\definecolor{purple}{RGB}{204, 204, 255} 
\definecolor{gray}{RGB}{229, 229, 229}
\usepackage{authblk}
  
\title{OPBO: Order-Preserving Bayesian Optimization}

\author[1]{Wei Peng}
\author[1,2]{Jianchen Hu*}
\author[2]{Kang Liu}
\author[1]{Qiaozhu Zhai}

\affil[1]{School of Automation Science and Engineering, Xi'an Jiaotong University, Xi'an, China}
\affil[2]{School of Future Technology, Xi'an Jiaotong University, Xi'an, China}

\begin{document}
\maketitle

\begin{abstract}
Bayesian optimization is an effective method for solving expensive black-box optimization problems. Most existing methods use Gaussian processes (GP) as the surrogate model for approximating the black-box objective function, it is well-known that it can fail in high-dimensional space (e.g., dimension over 500). We argue that the reliance of GP on precise numerical fitting is fundamentally ill-suited in high-dimensional space, where it leads to prohibitive computational complexity. In order to address this, we propose a simple order-preserving Bayesian optimization (OPBO) method, where the surrogate model preserves the \emph{order}, instead of the \emph{value}, of the black-box objective function. Then we can use a simple but effective OP neural network (NN) to replace GP as the surrogate model. Moreover, instead of searching for the best solution from the acquisition model, we select good-enough solutions in the ordinal set to reduce computational cost. The experimental results show that for high-dimensional (over 500) black-box optimization problems, the proposed OPBO significantly outperforms traditional BO methods based on regression NN and GP. The source code is available at \url{https://github.com/pengwei222/OPBO}.
\end{abstract}

\section{Introduction}\label{sec_intro}
Black-box optimization refers to the problem of searching for an input so that the output of a black-box system is optimized, the design can only rely on past input and output data collected from stimulating the system. In such problems, the objective function is considered a black-box and we have no access to its structure or gradient information. In reality, it is usually extremely expensive to collect rich enough data to describe the high-dimensional black-box objective function. For example, the design space exploration problem in chip design requires a timely and good enough parameter configuration with only limited tested data given (the data can be unreliable since they are collected from a cycle-accurate simulator) \cite{LZH:24}. The given dataset is only a small portion in the whole high-dimensional design space. Therefore, the main goal of black-box optimization is to find a solution close to the global optimum efficiently. The Bayesian optimization (BO) techniques for black-box functions have been developed in machine learning~\cite{Snoek2012PracticalBO, Dewancker2016BayesianOF}, with applications including hyperparameter tuning in deep learning~\cite{Snoek2012PracticalBO}, policy search in reinforcement learning~\cite{10.1007/s10472-015-9463-9}.

BO has proven to be a well-established approach for black-box optimization~\cite{frazier2018tutorialbayesianoptimization,7352306}. It iteratively implement the following two steps: 1) build a surrogate model to approximate the unknown objective function; 2) use an acquisition function to balance exploration and exploitation, guiding the selection of the next evaluation point. Gaussian processes (GP) model is widely used as a surrogate model in BO, due to its ability to provide accurate prediction and quantification of the model uncertainty. While GP performs well in low-dimensional problems (e.g., dimension less than 20), it often scale poorly to high-dimensional problems~\cite{frazier2018tutorialbayesianoptimization}. This problem arises due to several reasons. First, its reliance on precise numerical relationships for curve fitting becomes a bottleneck. In high-dimensional space, the curse of dimensionality renders data points exceedingly sparse, making accurate interpolation or pointwise fitting extremely difficult and prone to overfitting noise. Second, its computational complexity is dominated by kernel matrix construction. This cubic scaling with data dimensionality creates a severe computational bottleneck for high-dimensional problems. 

The modeling limitations and computational complexity of GP in high-dimensional space limits its practical application. In contrast, neural network (NN) exhibit stronger adaptability when handling high-dimensional data~\cite{DNGO, NeuralBO}. On one hand, NN possesses superior representational capacity for modeling high-dimensional function. On the other hand, unlike GP, the computational complexity of NN scales almost linearly with data size. Previous studies have attempted to apply deep neural networks (DNN) to optimization tasks, such as black-box optimization in continuous search space~\cite{PariaBeG} and contextual bandit problems in discrete space~\cite{10.5555/3524938.3526003}. However, most existing methods are based on Bayesian neural networks (BNN), whose posterior inference relies on computationally expensive sampling techniques~\cite{NIPS2016_a96d3afe}. More importantly, neither BNN nor DNN resolve the problem of precise numerical fitting in high-dimensional data. This leads to notable limitations in modeling. 

In order to overcome the modeling difficulty in high-dimensional space, we note that when selecting the next sampling point via an acquisition function in BO, we care more about the relative rank quality rather than the exact value of the candidate solution. This point aligns closely with ranking neural networks (RNN) in \cite{10.1145/1102351.1102363}. By directly modeling preference relationships between samples through comparative learning, the RNN becomes insensitive to the absolute scale of function value and it can focus on capturing the relative orders instead. This characteristic makes RNN particularly suitable for high-dimensional optimization problems. RNN is commonly used in areas such as recommendation systems and information retrieval, with the core task of generating relative rankings of items rather than predicting absolute scores. Another related application work preserving the candidate order is \cite{LZH:24}, which utilizes a transformation technique to preserve the order of the black-box function through a linear surrogate model for solving the design space exploration problem in processor design. 

Motivated by \cite{10.1145/1102351.1102363} and \cite{LZH:24}, we propose a general order-preserving (OP) surrogate model for BO in this work. The proposed model directly learns the ordinal relationships between samples, rather than their precise numerical values. This shift from value to order yields a representation that is more robust to uncertainty and noise. Furthermore, we relax the objective of the acquisition model from finding the optimal solution to selecting the good-enough solutions. This approach prioritizes practical acceptability over theoretical optimality. We conducted systematic experiments on benchmark high-dimensional (with dimension over 500) black-box problems. The results demonstrate that the proposed OP surrogate model significantly outperforms GP and regression NN in performance.

\section{Related Work}\label{sec_work}

BO is a key technique for solving black-box optimization problems~\cite{frazier2018tutorialbayesianoptimization,7352306}. The most commonly used surrogate model for BO is GP due to its flexibility in representing uncertainty and well-calibrated posterior distributions. The efficiency of the optimization process largely depends on the choice of acquisition function, such as Expected Improvement (EI)~\cite{ei} and Upper Confidence Bound (UCB)~\cite{ucb}. BO has achieved significant success in low-dimensional (with dimension less than 20) problems with limited sample budgets. However, Standard GP-based BO encounters notable challenges in high-dimensional space, particularly when dealing with non-stationary/noisy functions or scaling to large numbers of observations. On one hand, the number of observed points becomes sparse in high-dimensional space, making it difficult to accurately estimate the posterior distribution, which leads to a notable decline in the performance of both the surrogate and acquisition models. On the other hand, high dimensios exacerbates the increase in cumulative regret (i.e., the accumulated gap between the observed function values and the global optimum), resulting in overall performance degradation. 

In order to overcome the difficulties of high-dimensional optimization, one mainstream approach based on GP involves the use of random embedding strategy~\cite{10.5555/2540128.2540383,10.5555/3020751.3020776}. This method leverages the decomposable nature of the objective function, expressing the original high-dimensional function as a sum of several lower-dimensional sub-functions, under the assumption that each sub-function has a much lower intrinsic dimension. However, such approach requires training a large number of GP models, and its effectiveness depends on the mapping between high-dimensional space and the unknown low-dimensional subspaces, making it difficult to scale~\cite{10.5555/3020751.3020776,wang2016bayesianoptimizationbilliondimensions, hesbo}. HesBO~\cite{hesbo} extends GP-based BO algorithms to high-dimensional problems through a novel subspace embedding. This embedding effectively addresses the limitations inherent in the Gaussian projection methods employed by~\cite{wang2016bayesianoptimizationbilliondimensions,binois2015warpedkernelimprovingrobustness, 10.1007/s10898-019-00839-1}. Another class of method relies on search space constraint mechanism to improve algorithmic efficiency by restricting the searching region. For example, Thompson Sampling \cite{TS} randomly samples functions from the posterior distribution for optimization, effectively supports the evaluation of large batches of sample points. TuRBO \cite{TuRBO} combines adaptive trust regions with local optimization strategy, decomposing the global problem into multiple local optimization processes and thereby mitigating the curse of dimensionality.

Beyond random embedding method, another mainstream direction involves replacing the GP surrogate model with other functions. For instance, the work in \cite{SMAC} employed random forest as the surrogate model, while \cite{DNGO} used a linear regression network for feature extraction combined with a BNN for uncertainty estimation. In recent years, NN has been widely adopted as a substitute for GP as surrogate model in high-dimensional problems. For example, the work in \cite{NeuralBO} provided theoretical regret bounds for NN-based surrogate model using neural tangent kernel theory. However, such model often converges slowly in high-dimensional space and still requires a large number of samples. Another work called PFNs4BO \cite{PFNs4BO} employed a type of pre-trained tabular foundation model as a replacement for GP. Nevertheless, this model relies on prior data to train the NN, leading to significantly increased computational and memory costs.

In summary, finding a simple, flexible and powerful surrogate model in BO for high-dimensional black-box optimization problem is always a difficult task. We propose an OPBO method that includes an OP simple, flexible and powerful NN surrogate model. We focus on minimizing the more useful order approximation error, instead of the precise value approximation error, so that the surrogate model can be simple enough which requires greatly reduced sampling budge.

\section{Method}\label{sec_method}
\subsection{Bayesian Optimization}
The noisy black-box optimization problem is formulated as
\[
\text{Find } \mathbf{x}^* \in \Omega \text{ such that } f(\mathbf{x}^*) \leq f(\mathbf{x}),\ \forall \mathbf{x} \in \Omega;
\]
where $f: \Omega \to \mathbb{R}$ and $\Omega = [0, 1]^d$. The black-box function is accessed via noisy observations $y(\mathbf{x}) = f(\mathbf{x}) + \varepsilon$, where $\varepsilon \sim \mathcal{N}(0, \sigma^2)$. BO is an iterative procedure between modeling the objective function using a surrogate model $\hat{f}(y|\mathbf{x}, \mathcal{D}_r)$ based on all available observations $\mathcal{D}_r$ up to iteration $r$, and selecting the next query point by optimizing an acquisition function $\alpha(\hat{f}(y|\mathbf{x}, \mathcal{D}_r))$ derived from the surrogate model. Since each function evaluation is computationally expensive in black-box optimization, the acquisition function must balance exploring regions of high uncertainty with exploiting areas of known promise. 

GP has been widely adopted as a probabilistic surrogate model in BO due to its flexibility and analytical tractability. We present the pseudocode for the BO loop using GP in Algorithm~\ref{alg:bo_opt}. While GP performs well in low-dimensional space, it encounters several problems in high-dimensional settings. On one hand, the time complexity of GP is $O(n^3 + n^2 d)$, which increases with the dimension. On the other hand, the performance of GP deteriorates in high-dimensional space due to the data's complex, discrete structures, which deviates from the joint Gaussian assumption.

\begin{algorithm}
\caption{BO with GP}
\label{alg:bo_opt}
\begin{algorithmic}[1]
\Require Initial dataset $\mathcal{D}_0 = \{(\mathbf{x}_1, y_1), \ldots, (\mathbf{x}_k, y_k)\}$; initial size $k$; search space $\mathcal{X}$; total iterations $R$; black-box function $f$; sample size $N$ for Sampling function; acquisition function $\alpha$.
\State $\mathcal{D} \gets \mathcal{D}_0$ 
\For{$r = 1$ to $R$}
    \State Fit GP model $\hat{f}$ on data $\mathcal{D}$
    \State Generate candidate set $\mathcal{X}_{\text{cand}} \gets \text{Sampling}(\mathcal{X}, N)$
    \State Suggest $\mathbf{x} \in \arg \max_{\hat{\mathbf{x}} \in \mathcal{X}_{\text{cand}}} \alpha(\hat{\mathbf{x}}, \mathcal{D}, \hat{f})$ 
    \State Evaluate $\mathbf{x}$ on $f$ to obtain $y = f(\mathbf{x})$ 
    \State Update data: $\mathcal{D} \leftarrow \mathcal{D} \cup \{(\mathbf{x}, y)\}$
\EndFor
\State \Return $\mathbf{x}^* \gets \arg\min_{(\mathbf{x}_i, y_i) \in \mathcal{D}} y_i$ 
\end{algorithmic}
\end{algorithm}

\subsection{Order-Preserving Surrogate Objective}

In black-box optimization, the objective function often lacks analytical form, and real-world observations are usually contaminated by noise. Since ordinal relationships are generally easier to model and more robust to noise than exact values, the relative ordering between observations is often more critical than their absolute values. Therefore, we use OP function to formalize the order relationship between two functions, similar to \cite{LZH:24}, as defined below:

\begin{definition}[Order-Preserving Functions]
Two functions $f$ and $h$ with the same domain are said to be \textit{order-preserving} (denoted as $f\stackrel{OP}{\Longleftrightarrow} g$) if, for all $a, b$ in their domains, $f(a) \geq f(b)$ if and only if $h(a) \geq h(b)$.
\end{definition}

The concept of order-preserving describes when two different functions preserve the same ranking of elements in their domain. A sufficient condition for strict order-preserving is that one function is a non-decreasing transformation of the other, i.e., $h = g \circ f$ for some non-decreasing function $g$. In such a case, $f$ and $h$ preserve the same order relation. In order to ground this concept, consider the two-dimensional Gaussian Radial Basis Functions (RBF) $F(x_1, x_2) = e^{-\frac{x_1^2 + x_2^2}{2\sigma^2}}$ and a corresponding OP function $H(x_1, x_2) = -(x_1^2 + x_2^2)$, with $\sigma$ a constant. Observing that $H(x_1, x_2) = 2\ln(F(x_1, x_2))$, and since the logarithm is strictly increasing, it follows immediately that $F\stackrel{OP}{\Longleftrightarrow}H$. The OP property guarantees that the contour profiles of the functions remain consistent, undergoing only a scaling transformation.

\textbf{Order-Preserving Analysis:} In order to analyze the OP property, we employ two two-dimensional true functions: \( f(\mathbf{x}) = e^{-\frac{x_1^2 + x_2^2}{2}} \) and \( f(\mathbf{x}) = e^{\frac{x_1^2 + x_2^2}{2}} \), both defined on the domain \([-6, 6]\). In order to mitigate the effects of distributional differences, the function values were normalized to a Gaussian distribution with $\mu=0$ and $\sigma=1$ using the transformation \( f(\mathbf{x}) = \frac{f(\mathbf{x}) - \mu}{\sigma} \). Since $h(\mathbf{x})=-(x_1^2 + x_2^2)$ and $h(\mathbf{x})=(x_1^2 + x_2^2)$ are known OP functions for the original true functions, we call $h(\mathbf{x})$ the OP function. In contrast, the proposed OP surrogate model (a two-layer MLP structure as in \cite{Rumelhart1986LearningRB}, optimized with the loss in Equation \eqref{eq4})  in this paper is called OP model.  

\begin{figure*}
\centering
\begin{subfigure}[b]{1.0\linewidth}
    \centering
    \includegraphics[width=\textwidth]{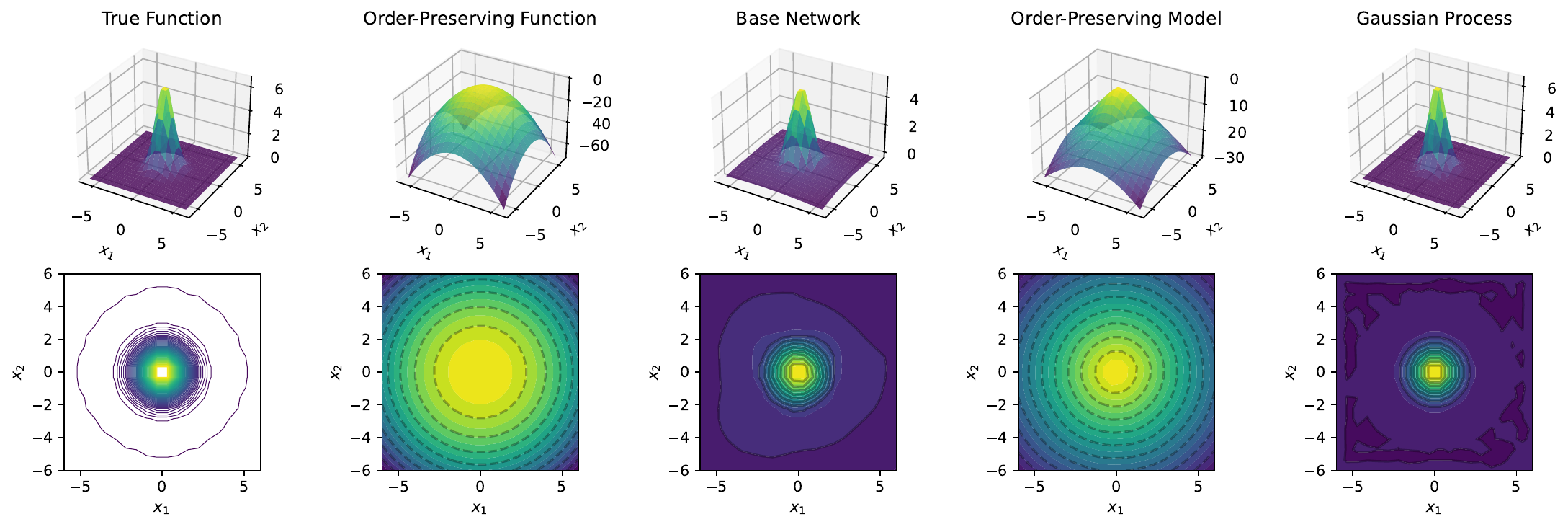}
    \caption{Comparison of true function $f(\mathbf{x}) = e^{-\frac{x_1^2 + x_2^2}{2}}$, OP function $h(\mathbf{x}) = -{(x_1^2 + x_2^2)}$, base NN, OP model \eqref{eq4}, GP model}
    \label{fig:con1}
\end{subfigure}

\vspace{0.3cm} 

\begin{subfigure}[b]{1.0\linewidth}
    \centering
    \includegraphics[width=\textwidth]{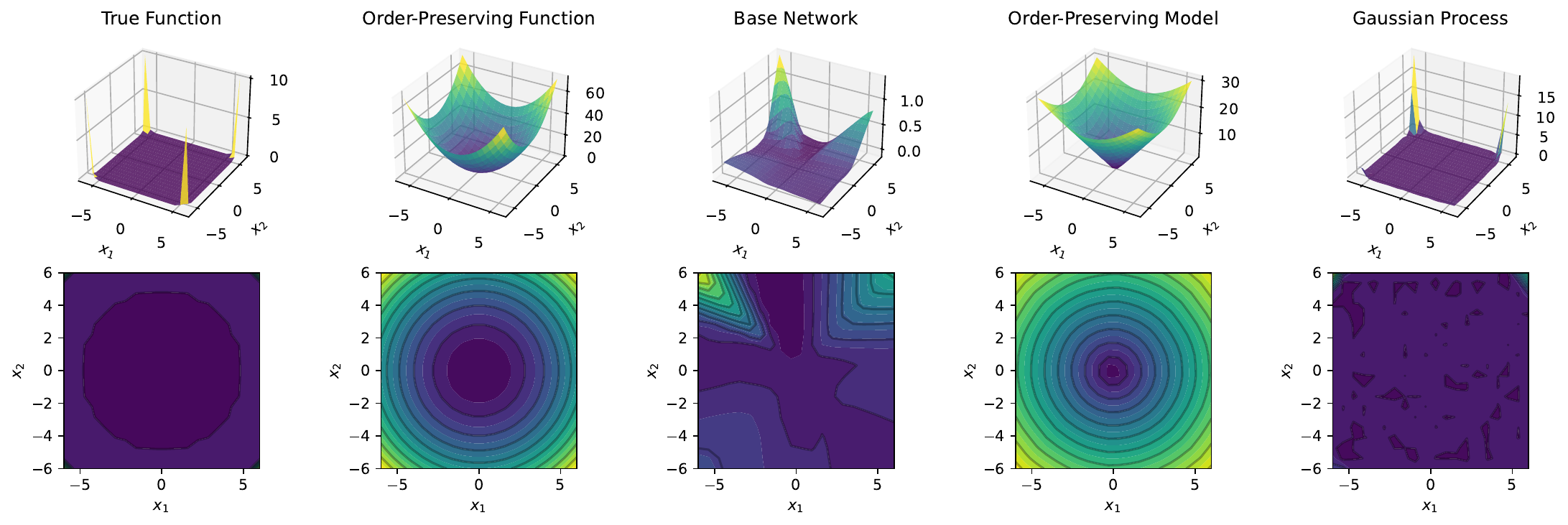}
    \caption{Comparison of true function $f(\mathbf{x}) = e^{\frac{x_1^2 + x_2^2}{2}}$, OP function $h(\mathbf{x}) = {(x_1^2 + x_2^2)}$, base NN, OP model \eqref{eq4}, GP model}
    \label{fig:con2}
\end{subfigure}

\caption{Analysis of OP properties for different surrogate models.}
\label{fig:oo_results}
\end{figure*}

As shown in Figure~\ref{fig:oo_results}, we compare the true function $f(\mathbf{x})$ with the OP function $h(\mathbf{x})$, based NN model in \cite{Rumelhart1986LearningRB}, OP model \eqref{eq4}, and GP model in \cite{ei}. These surrogate models are applied to fit the true function and its contour surfaces, respectively. Figure~\ref{fig:con1} illustrates the contour surfaces of the RBF function \( f(\mathbf{x}) = e^{-\frac{x_1^2 + x_2^2}{2}} \) , which is characterized by a single central peak surrounded by flat regions. It can be observed that GP fits well in the peak region, but it fits poorly in the flat areas due to the deviation of the data from a Gaussian distribution. NN (a two-layer MLP structure as in \cite{Rumelhart1986LearningRB}, optimized with the MSE loss) also fits poorly in the flat regions. In contrast, both the constructed OP function $h(\mathbf{x})$ (already known to be OP) and the proposed general OP model (will shown to be OP later) accurately capture the contour lines of the function. Figure~\ref{fig:con2} displays the contour surfaces of the function \( f(\mathbf{x}) = e^{\frac{x_1^2 + x_2^2}{2}} \), which features peaks at the periphery and a flat central region. Here, the GP model fits the flat central region reasonably well but fails to capture the peaks, as the data in those regions do not conform to a Gaussian distribution. NN performs even worse, completely fails to reflect the ordinal characteristics of the function. On the other hand, both the OP function and the OP model successfully and accurately represent the contour lines of the function. 

In order to evaluate the OP capability of the models, we employ Spearman’s rank coefficient of correlation. This nonparametric statistic measures the strength and direction of monotonic association between two ranked variables as
\begin{equation}
\rho = 1 - \frac{6\sum d_i^2}{n(n^2-1)}
\end{equation}
where $d_i$ represents the difference in ranks for the $i$-th observation, and $n$ denotes the total number of observations. The coefficient produces values in the range $[-1, 1]$, with $\rho = 1$ indicating perfect positive monotonic correlation, $\rho = -1$ indicating perfect negative monotonic correlation, and $\rho = 0$ indicating no monotonic relationship.

\begin{figure*}
\centering
\begin{subfigure}[b]{0.46\linewidth}
    \centering
    \includegraphics[width=\textwidth]{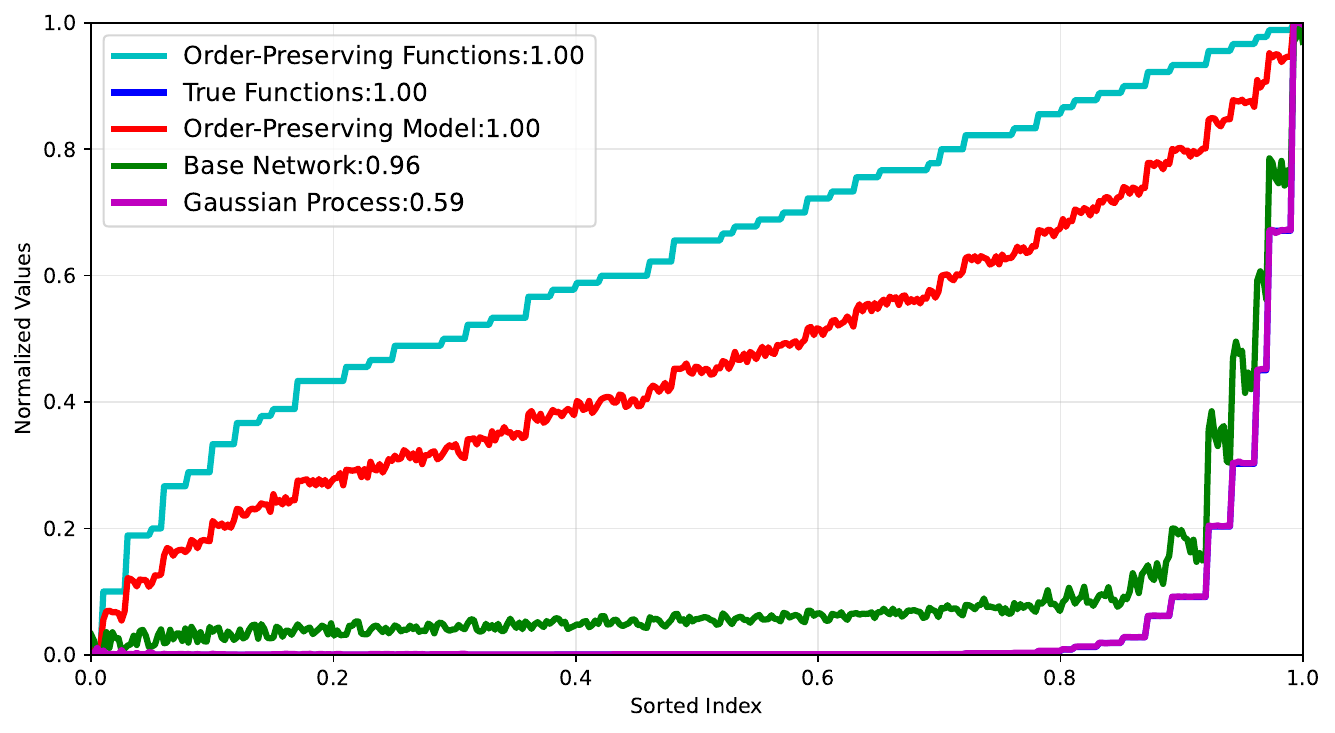}
    \caption{Comparison of OP function $h(\mathbf{x}) = -{(x_1^2 + x_2^2)}$, true function $f(\mathbf{x}) = e^{-\frac{x_1^2 + x_2^2}{2}}$, OP model, base NN, GP model}
    \label{fig:opc1}
\end{subfigure}
\hspace{0.05\linewidth} 
\begin{subfigure}[b]{0.46\linewidth}
    \centering
    \includegraphics[width=\textwidth]{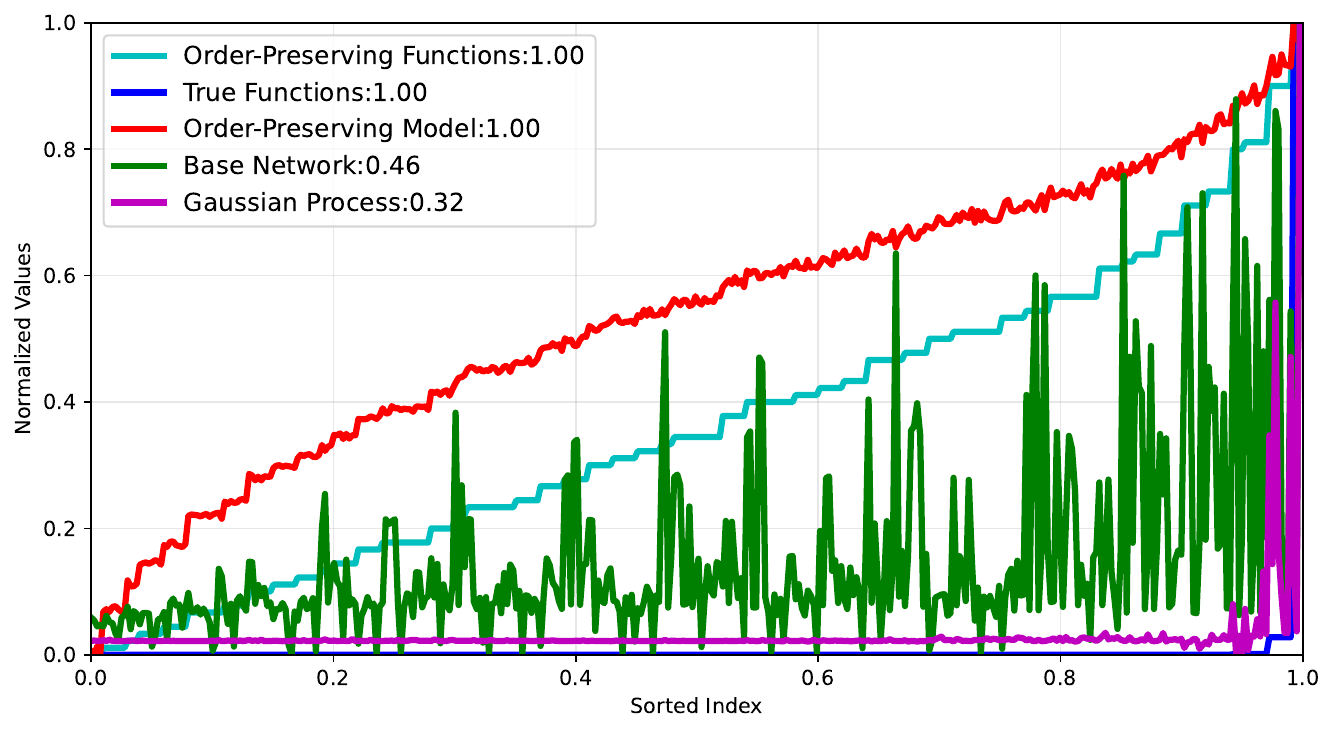}
    \caption{Comparison of OP function $h(\mathbf{x}) = {(x_1^2 + x_2^2)}$, true function $f(\mathbf{x}) = e^{\frac{x_1^2 + x_2^2}{2}}$, OP model, base NN, GP model}
    \label{fig:opc2}
\end{subfigure}

\caption{Comparison of OPC for different surrogate models.}
\label{fig:opc_results}
\end{figure*}

Figure~\ref{fig:opc_results} displays the ordered preserving curve (OPC) curves of the four models, with the Spearman correlation coefficient used to evaluate the ordinal relationships. The OPC (originally introduced for ordinal optimization in \cite{oo}) is a monotonically increasing performance curve with ordered collected design points. It represents the basic order structure of a problem. From Figure~\ref{fig:opc1}, when the data is subject to a highly non-uniform distribution, with many higher rank values clustered in a narrow low range (e.g., [0, 0.05]), the GP model's OPC aligns well with the true curve, yet its Spearman’s coefficient is low. This demonstrates that the GP surrogate is effective at modeling the overall functional trend, but it performs poorly in preserving the pairwise ordering among the densely packed low-performance points. From Figure~\ref{fig:opc2}, it can be observed that the GP model performs well when the distribution of the true function is uniform. However, the fitting performance of a GP deteriorates when the true function's distribution is highly uneven, with solutions concentrated in disparate high- or low-performance regions. The NN exhibits even more oscillatory. In contrast, the OP model effectively simplifies the representation while successfully capturing the underlying ordinal structure of the function. 

\subsection{Ordinal Optimization}

In many practical applications, particularly in optimization, a crude OP function can be sufficient. Two functions are considered coarsely OP if they share the same long-term monotonic trends. A canonical example is \( f(x) = \sin(20x) + 5x \) and its crude OP function \( h(x) = 5x \). While \( f \) exhibits high-frequency oscillations, its monotonic behavior is dominated by the linear term  \( 5x \). Consequently, for the purpose of optimization, the simpler function \( h \) can effectively serve as a surrogate for \( f \), significantly simplifying the problem.

\begin{figure*}
\centering
\begin{subfigure}[b]{0.17\linewidth}
    \centering
    \includegraphics[width=\textwidth]{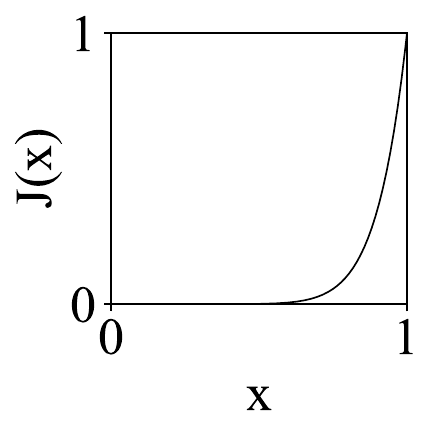}
    \caption{Flat}
    \label{fig:func1}
\end{subfigure}
\hfill
\begin{subfigure}[b]{0.17\linewidth}
    \centering
    \includegraphics[width=\textwidth]{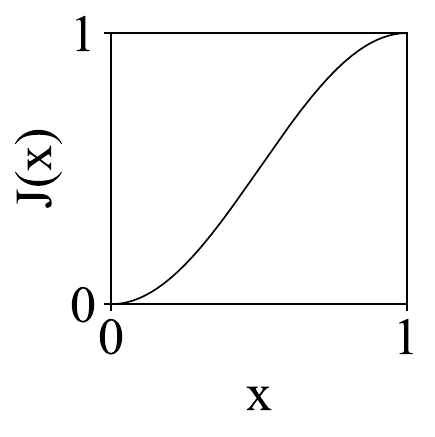}
    \caption{U-Shaped}
    \label{fig:func2}
\end{subfigure}
\hfill
\begin{subfigure}[b]{0.17\linewidth}
    \centering
    \includegraphics[width=\textwidth]{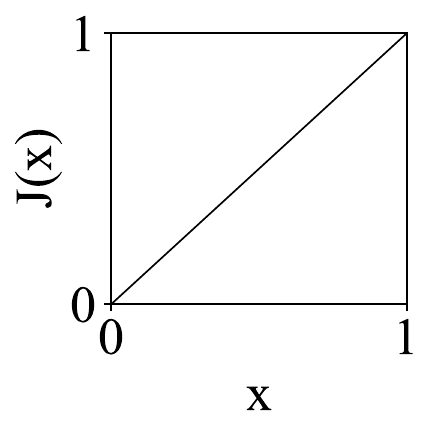}
    \caption{Neutral}
    \label{fig:func3}
\end{subfigure}
\hfill
\begin{subfigure}[b]{0.17\linewidth}
    \centering
    \includegraphics[width=\textwidth]{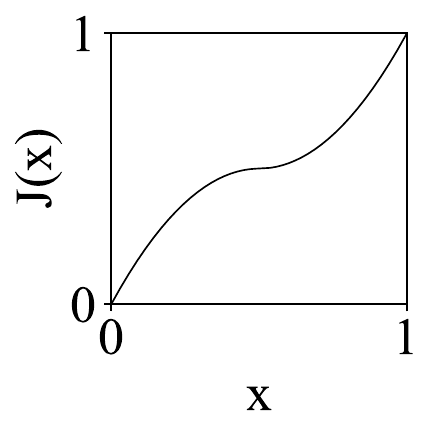}
    \caption{Bell}
    \label{fig:func4}
\end{subfigure}
\hfill
\begin{subfigure}[b]{0.17\linewidth}
    \centering
    \includegraphics[width=\textwidth]{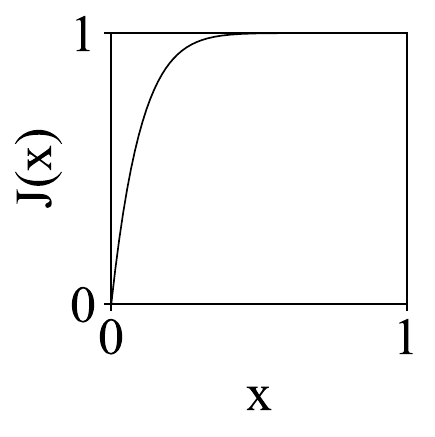}
    \caption{Steep}
    \label{fig:func5}
\end{subfigure}
\caption{Normalized OPCs.}
\label{fig:opc_type}
\end{figure*}

\begin{figure*}
\begin{subfigure}[b]{0.17\linewidth}
    \centering
    \includegraphics[width=\textwidth]{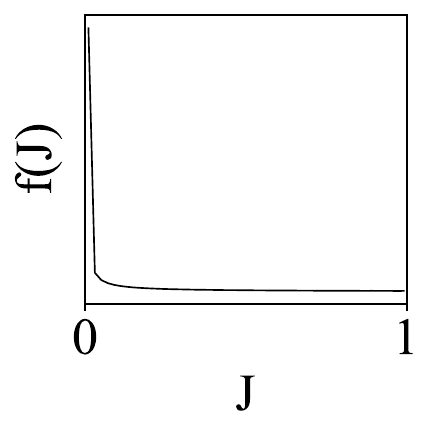}
    \caption{Flat}
    \label{fig:dens1}
\end{subfigure}
\hfill
\begin{subfigure}[b]{0.17\linewidth}
    \centering
    \includegraphics[width=\textwidth]{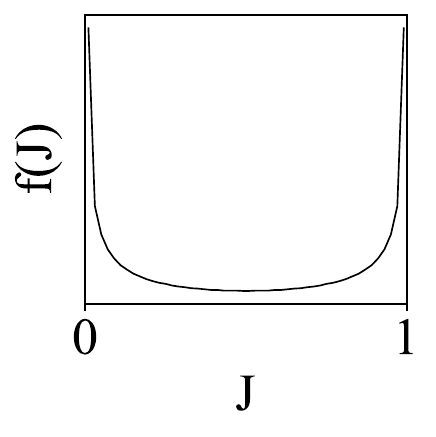}
    \caption{U-Shaped}
    \label{fig:dens2}
\end{subfigure}
\hfill
\begin{subfigure}[b]{0.17\linewidth}
    \centering
    \includegraphics[width=\textwidth]{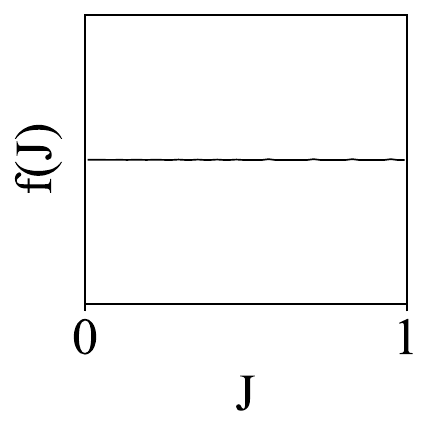}
    \caption{Neutral}
    \label{fig:dens3}
\end{subfigure}
\hfill
\begin{subfigure}[b]{0.17\linewidth}
    \centering
    \includegraphics[width=\textwidth]{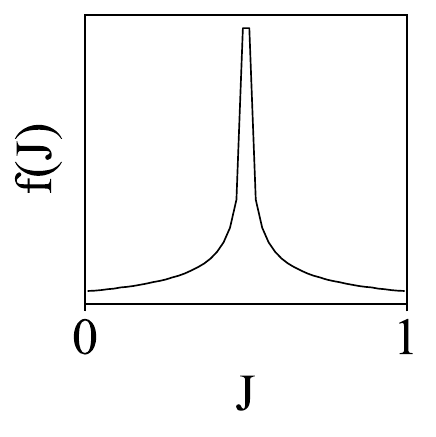}
    \caption{Bell}
    \label{fig:dens4}
\end{subfigure}
\hfill
\begin{subfigure}[b]{0.17\linewidth}
    \centering
    \includegraphics[width=\textwidth]{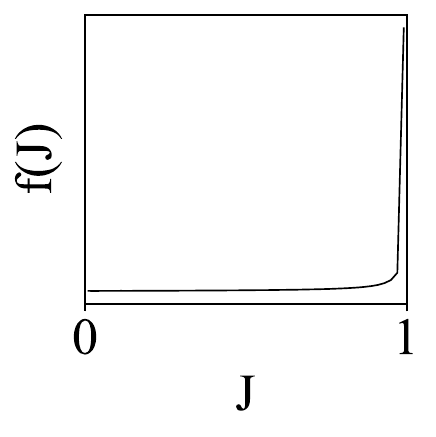}
    \caption{Steep}
    \label{fig:dens5}
\end{subfigure}

\caption{Normalized performance density functions.}
\label{fig:distribution_density}
\end{figure*}

\begin{definition}[Ordered Performance Curve]
Let \( J_{[1]} \geq J_{[2]} \geq \dots \geq J_{[N]} \) be a sequence of performance values sampled from the search space $\mathcal{X}$ sorted in ascending order. The OPC is defined by the points \( (x_i, y_i) \) for \( i = 1, \dots, N \), where:
\[
x_i = \frac{i-1}{N-1}, \quad y_i = -\frac{J_{[i]} - J_{[1]}}{J_{[N]} - J_{[1]}}.
\]
\end{definition}
Here, \( x_i \) represents the normalized rank, and \( y_i \) represents the normalized \( J_{[i]} \). As shown in Figure~\ref{fig:opc_type}, the OPC can be categorized into five types \cite{oo}: Flat (Fig.~\ref{fig:func1}), U-Shaped (Fig.~\ref{fig:func2}), Neutral (Fig.~\ref{fig:func3}), Bell (Fig.~\ref{fig:func4}), and Steep (Fig.~\ref{fig:func5}). As shown in Figure~\ref{fig:distribution_density}, different OPC type is characterized by a distinct distribution of inter-point distances: for Flat (Fig.~\ref{fig:dens1}), good solutions are clustered while poor ones are dispersed; for Steep (Fig.~\ref{fig:dens5}), poor solutions are clustered while good ones are scattered; for U-shaped (Fig.~\ref{fig:dens2}), intermediate solutions are spread out; and for Bell (Fig.~\ref{fig:dens4}), intermediate solutions are closely grouped; for the Neutral type (Fig.~\ref{fig:dens3}), the distances are uniformly distributed. The OPC type varies across different functions, and differ even between OP functions. The Neutral type, characterized by its smooth distribution, represents the simplest order relationship. Consequently, transforming the other four OPC types into the Neutral type would be highly beneficial for our modeling.

Therefore, we propose a general OP surrogate model. Instead of modeling scalar function values, the model now directly learns to predict rankings as
\begin{equation}
\theta^* = \arg\max_{\theta} \prod_{i=1}^{N} p(\pi_i \mid \mathbf{X}_{i-1}, \theta) \label{eq2}
\end{equation}
where \( p(\pi_i \mid \mathbf{X}_{i-1}, \theta) \) is the probability of the ranking \( \pi_i \) conditioned on the prior set \( \mathbf{X}_{i-1} \). By assuming a uniform probability over permutations $\pi_i$, we effectively simplify the OPC towards the Neutral type, thereby promoting a smoother and more tractable modeling order relationship.

In black-box optimization, finding the global optimum is often costly. Consequently, we shift our goal from finding the best solution to acquiring a ``good-enough'' solution, defined as any designs ranked in the top-$g$. This leads to a more robust strategy: training a single model to predict the entire top-$g$ set, rather than combining the best outputs from an ensemble of $g$ separate models. The approach significantly reduces computational costs and enhances the success probability of the single model's predictions.

Algorithm~\ref{alg:oo_opt} details the procedure for our OPBO algorithm. $R$ represents the total number of iteration rounds, which is determined by the simulation budget. $N$ represents the candidate sample size, which is set to $\text{dim} \times 10$ to ensure adequate coverage of the high-dimensional search space, as a sparse distribution of points cannot guarantee the inclusion of near-optimal solutions. $\alpha$ represents the acquisition function (common acquisition functions are shown in Table ~\ref{tab:acquisition_functions}). $g$ represents the size of the good-enough set (we set $g=10$ in this work). $\text{Sampling}(\mathcal{X}, N)$ is designed to sample $N$ candidate points from the search space $\mathcal{X}$ (common sampling strategies are shown in Table ~\ref{tab:sampling_methods}).

\begin{algorithm}
\caption{The proposed OPBO Algorithm}
\label{alg:oo_opt}
\begin{algorithmic}[1]
\Require Initial dataset $\mathcal{D}_0 = \{(\mathbf{x}_1, y_1), \ldots, (\mathbf{x}_k, y_k)\}$; initial size $k$; search space $\mathcal{X}$; total iterations $R$; black-box function $f$; sample size $N$ for Sampling function; acquisition function $\alpha$; good-enough set $\mathcal{G}$; $\mathcal{G}$ size $g$.
\State $\mathcal{D} \gets \mathcal{D}_0$
\For{$r = 1$ to $R$}
    \State Fit OP surrogate model $\hat{f}$ on $\mathcal{D}$
    \State Generate candidate set $\mathcal{X}_{\text{cand}} \gets \text{Sampling}(\mathcal{X}, N)$
    \State Select good-enough set $\mathcal{G} \gets \operatorname{top-}g\max_{\mathbf{x} \in \mathcal{X}_{\text{cand}}} \alpha(\hat{\mathbf{x}}, \mathcal{D}, \hat{f})$ 
    \State Evaluate good-enough set $\mathcal{G}$ on $f$ to obtain $y_j = f(\mathbf{x}_j)$ for each $\mathbf{x}_j \in \mathcal{G}$
    \State Update data: $\mathcal{D} \leftarrow \mathcal{D} \cup \{(\mathbf{x}_j, y_j) : \mathbf{x}_j \in \mathcal{G}\}$
\EndFor
\State \Return $\mathbf{x}^* \gets \arg\min_{(\mathbf{x}_i, y_i) \in \mathcal{D}} y_i$
\end{algorithmic}
\end{algorithm}

\begin{table}[h]
\centering
\caption{Common acquisition functions in BO}
\begin{tabular}{llp{5.66cm}}
\toprule
\textbf{Acquisition Function} & \textbf{Mathematical Formulation} & \textbf{Description} \\
\midrule
\makecell[l]{Expected Improvement\\(EI) \cite{ei}} & $\alpha_{\text{EI}}(\mathbf{x}) = \mathbb{E}[\max(f(\mathbf{x}) - f(\mathbf{x}^+), 0)]$ & $f(\mathbf{x}^+)$ is the current best observed value. Maximizes the expected improvement over the current optimum. \\
\makecell[l]{Upper Confidence \\Bound (UCB) \cite{ucb}}   & $\alpha_{\text{UCB}}(\mathbf{x}) = \mu(\mathbf{x}) + \kappa \sigma(\mathbf{x})$ & $\mu(\mathbf{x})$ and $\sigma(\mathbf{x})$ are the posterior mean and standard deviation. $\kappa \geq 0$ controls the trade-off between exploration $\sigma(\mathbf{x})$ and exploitation $\mu(\mathbf{x})$. \\
\makecell[l]{Thompson Sampling\\(TS) \cite{TS}} & $\begin{aligned} &\text{Sample a function: } \hat{f} \sim \mathcal{GP}(\mu(\mathbf{x}), \kappa(\mathbf{x}, \mathbf{x}')), \\ &\text{then select: } \mathbf{x}_{\text{next}} = \arg\max_{\mathbf{x}} \hat{f}(\mathbf{x}). \end{aligned}$ & A probabilistic strategy where $\mu(\mathbf{x})$ is the posterior mean function and $\kappa(\mathbf{x}, \mathbf{x}')$ is the kernel function defining the covariance structure. \\
\bottomrule
\end{tabular}
\label{tab:acquisition_functions}
\end{table}

\begin{table}[h]
\centering
\caption{Common sampling strategies for candidate generation}
\begin{tabular}{lp{10.2cm}}
\toprule
\textbf{Sampling Method} & \textbf{Description}  \\
\midrule
Random Sampling (RS) \cite{rs}& Uniformly random points within the bounds of $\mathcal{X}$.  \\
Latin Hypercube Sampling (LHS) \cite{lhs}& Stratified sampling ensuring each dimension is divided into $N$ equal intervals with exactly one sample per interval. \\
Grid Sampling (GS) \cite{gird} & Regular grid over the search space (e.g., equally spaced points along each dimension). \\
\bottomrule
\end{tabular}
\label{tab:sampling_methods}
\end{table}

Compared to BO, OPBO shifting the focus from precise cardinal values to robust ordinal relationships, which can be easier to model, especially in complex or noisy scenarios. This shift not only enhances data efficiency but also naturally softens the optimization objective; by targeting a ``good-enough'' set $\mathcal{G}$ rather than a single best point, the method mitigates the risk of over-exploitation and aligns more closely with the practical goal of finding a satisfactory solution under limited budget.

\subsection{Model Training}

We train the OP NN by minimizing the OP surrogate objective, which is defined as the negative log-likelihood of the observed rankings. Given the predicted scores $\mathbf{s} = [s_1, s_2, \dots, s_n]^\top$ and the corresponding true function values $\mathbf{y} = [y_1, y_2, \dots, y_n]^\top$ for a batch of samples, the samples are first sorted in descending order according to the true function values to obtain the permutation indices $\pi$. The predicted scores are then reordered according to this permutation, and the conditional probability at each position is computed. Specifically, given the first $i-1$ already-selected samples, the probability of selecting the $i$-th sample is defined as:

\begin{equation}
P(i \mid {1, 2, \dots, i-1}) = \frac{\exp(s_{\pi(i)})}{\sum_{k=i}^n \exp(s_{\pi(k)})}
\end{equation}

The likelihood of the entire permutation is the product of all such conditional probabilities. The model parameters are optimized by minimizing the negative log-likelihood:

\begin{equation}
\mathcal{L} = -\sum_{i=1}^n \log P(i \mid {1, 2, \dots, i-1}) = -\sum_{i=1}^n \left[ s_{\pi(i)} - \log\left( \sum_{k=i}^n \exp(s_{\pi(k)}) \right) \right] \label{eq4}
\end{equation}

This loss function directly optimizes the generation probability of the full permutation in an end-to-end manner, effectively capturing order relationships among samples.

\begin{remark}
We did not discuss the design of acquisition function or constraints in this work. Even though we apply the EI acquisition function and Thompson sampling for consistency with previous works, other acquisition functions and sampling methoeds are also applicable. The constraints can also be adopted directly in our framework. In fact, in discrete high-dimensional design space, one can apply our proposed OP modeling method to approximate black-box constraints (often appear in robot planning in unknown environment). This can be an interesting topic and it is beyond the scope of this work. 
\end{remark}

\section{Experiments}\label{sec_result}

In this section, we evaluate the effectiveness of the proposed OP surrogate model for black-box optimization tasks across a range of synthetic benchmark functions.

\subsection{Experimental Setup}
Across all experiments, we benchmarked the OPBO against commonly used surrogate models in black-box optimization, including GP and DNN. The performance was evaluated within four classical optimization frameworks: Standard BO~\cite{BO_TS}, TuRBO~\cite{TuRBO}, HEBO~\cite{hebo}, HesBO~\cite{hesbo}.

\textbf{Test Problems} As shown in Table \ref{tab:benchmark_funcs}, this study employs a suite of four classical synthetic optimization problems: the Ackley, Levy, Rosenbrock, and DixonPrice functions, to evaluate algorithm performance across diverse high-dimensional scenarios. The mathematical definitions of these functions can be found in the following authoritative optimization test suite: \url{http://www.sfu.ca/~ssurjano/optimization.html}. The domain for all functions is uniformly set to $[-5, 10]^d$, where $d$ denotes the dimensionality. Tests are conducted across a dimensional range of $d \in \{600, 700, 800, 900, 1000\}$, resulting in a total of $5 \times 4 = 20$ distinct problem variants. All benchmark functions are formulated as minimization tasks to facilitate a precise evaluation of the algorithms' convergence performance and solution accuracy.

\begin{table}[htbp]
    \centering
    \caption{High-dimensional benchmark problems.}
    \label{tab:benchmark_funcs}
    \begin{tabular}{@{}p{2cm} p{13.5cm}@{}}
        \toprule
        \textbf{Function} & \textbf{Mathematical Definition} \\
        \midrule
        Ackley &
        \makecell[l]{
            $f(\mathbf{x}) = -a \exp\Bigl(-b \sqrt{\frac{1}{d}\sum_{i=1}^{d} x_i^2}\Bigr) 
            - \exp\Bigl(\frac{1}{d}\sum_{i=1}^{d} \cos(c x_i)\Bigr) + a + \exp(1)$ \\
            where $a = 20$, $b = 0.2$, $c = 2\pi$.
        } \\
        Levy &
        \makecell[l]{
            $f(\mathbf{x}) = \sin^2(\pi w_1) + \sum_{i=1}^{d-1} (w_i - 1)^2 
            \bigl[1 + 10 \sin^2(\pi w_i + 1)\bigr] + (w_d - 1)^2 \bigl[1 + \sin^2(2\pi w_d)\bigr]$ \\
            where $w_i = 1 + \frac{x_i - 1}{4}$, $\forall i=1,\dots,d$.
        } \\
        Rosenbrock &
        $f(\mathbf{x}) = \sum_{i=1}^{d-1} \bigl[100(x_{i+1} - x_i^2)^2 + (x_i - 1)^2\bigr]$.\\
        Dixon-Price &
        $f(\mathbf{x}) = (x_1 - 1)^2 + \sum_{i=2}^{d} i \bigl(2x_i^2 - x_{i-1}\bigr)^2$. \\
        \bottomrule
    \end{tabular}
\end{table}

\textbf{Baseline Methods and Implementation Details} In this experiment, the OPBO was integrated into four optimization frameworks (Standard BO~\cite{BO_TS}, TuRBO~\cite{TuRBO}, HEBO~\cite{hebo}, HesBO~\cite{hesbo}) replacing the original Surrogate Model (GP, NN). The implementation details are as follows:

\begin{itemize}
\item Standard BO~\cite{BO_TS}: We implement the Standard BO framework using a GP surrogate model with the Squared Exponential (SE) kernel, and the Thompson Sampling (TS) acquisition function.
\item TuRBO~\cite{TuRBO}: TuRBO operates by fitting multiple local models and dynamically allocating samples among them through a principled, bandit-based global strategy. We employ the original implementation at \url{https://github.com/uber-research/TuRBO}.
\item HEBO~\cite{hebo}: HEBO handles complex noise processes via input warping and output transformations while enabling multi-objective acquisition optimization through evolutionary Pareto-frontier search. We employ the original implementation at \url{https://github.com/huawei-noah/HEBO}.
\item HesBO~\cite{hesbo}: HesBO performs high-dimensional BO by projecting the search space onto a low-dimensional subspace with bounded error. We employ the original implementation at \url{https://github.com/aminnayebi/HesBO}.
\end{itemize}

\textbf{Algorithm Tests} For each test problem, we conducted 10 independent trials using different random seeds to ensure statistical significance. For a fair comparison, all algorithms were initialized with the same dataset of 10 points generated via Latin Hypercube Sampling (LHS), employing a consistent random seed strategy. During each iteration, every algorithm selected the next sample point for evaluation based on its acquisition function and obtained the true function value at that point to update its surrogate model. All hyperparameters for both the NN and RankNet (Order-Preserving) models were kept identical. We employed a two-layer neural network architecture with 128 neurons in both the hidden layer and the output layer. The weights of the hidden layers were initialized using the Xavier method. The models were trained with the Adam optimizer, using a batch size of 2000, for 50 epochs, and with a learning rate of 0.01.

\subsection{Evaluation Metrics}

\textbf{Optimization Fixed-budget Convergence Analysis} Fixed-budget evaluation is a widely adopted technique for comparing the efficiency of optimization algorithms by allocating predetermined computational resources for their execution. In our study, we employ a "fixed-iteration" approach to ensure
all algorithms are allocated approximately equal computational time budgets. Therefore we execute TuRBO~\cite{TuRBO} and BO~\cite{BO_TS} for 500 iterations while running HesBO~\cite{hesbo} for 200 iterations and HEBO~\cite{hebo} for 110 iterations with the first $dim/10$ iterations used for sampling. This design enables a direct assessment of convergence efficiency under equivalent resource constraints, providing a realistic measure of each method's practical optimization capability.

\textbf{Statistical Ranking} For comprehensively compare and evaluate the performance of the BO algorithms, statistical ranking techniques are employed instead of direct performance measurements of the optimization outcome. In this study, we define the optimization performance result as the median of the minimum result found (final incumbent) across the 10 optimization trials of each algorithm. By statistically ranking the results, we were able to standardize the comparisons across different problems, since various optimization challenges can produce objective values of vastly different magnitudes. Furthermore, using this ranking allowed us to reduce the distorting effects of unusual or extreme data points that might influence our evaluation.
We conduct our statistical analysis using the Friedman and Wilcoxon signed-rank tests, complemented by Holm’s alpha correction. These non-parametric approaches excel at processing benchmarking result data without assuming specific distributions, which is critical for handling optimization results with outliers. These statistical methods effectively handle the dependencies in our setup, where we used the same initial samples and seeds to test all algorithms. The Wilcoxon signed-rank test addresses paired comparisons between algorithms, while the Friedman test manages problem-specific grouping effects. For multiple algorithm comparisons, we used Holm’s alpha correction to control error rates.

\textbf{Algorithm Runtime Recording} Each algorithm is timed for the total time it takes to run one experiment trial. We take the overall mean over the 10 trials and over all benchmark problems, resulting in a single value average time (tavg) to represent the average runtime of each algorithm.

\subsection{Results and Evaluations}

\begin{figure*}
\centering
\begin{subfigure}[b]{0.6\linewidth}
    \centering
    \includegraphics[width=\textwidth]{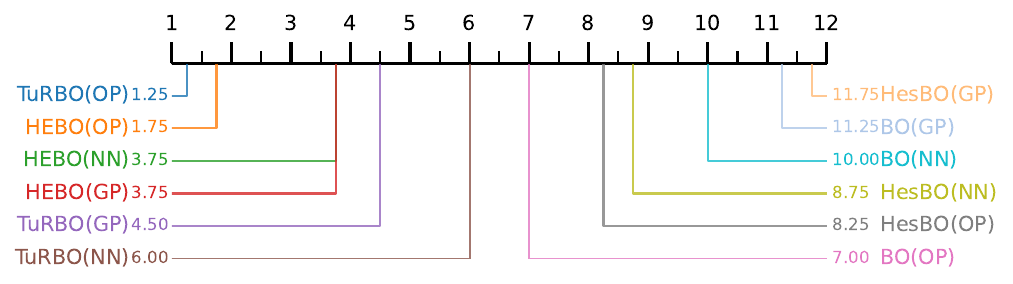}
    \caption{Algorithm Rank}
    \label{fig:rank}
\end{subfigure}
\hspace{0.05\linewidth} 
\begin{subfigure}[b]{0.4\linewidth}
    \centering
    \includegraphics[width=\textwidth]{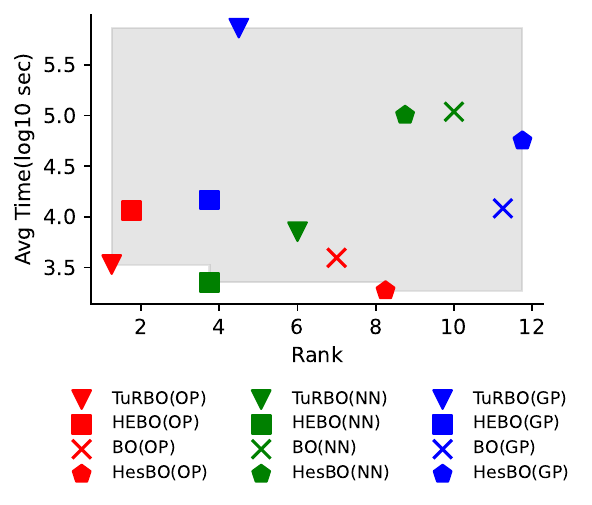}
    \caption{Pareto Froniter Plot}
    \label{fig:pareto}
\end{subfigure}

\caption{(a): Statistical ranking of the overall performance across all benchmark problems.
TuRBO(OP) ranked the top at optimization results. (b): Plot of average time vs overall statistical rank. The shorter time and smaller rank perform better (bottom left corner), so we show TuRBO(OP) at the Paerto front as the best algorithm.}
\end{figure*}

\textbf{Overall Statistical Ranking and Algorithm Runtime Tradeoffs} Across all problem variants, Figure~\ref{fig:rank} shows that TuRBO(OP) achieves the top statistical performance rank based on the optimization results (1.25), followed by HEBO(OP) (1.75), HEBO(NN) (3.75). Figure~\ref{fig:pareto} plots the average run time tavg of each algorithm versus the statistical rank. Due to the parallelism of neural networks, the time required by neural networks is significantly shorter than that by GP. TuRBO(OP) requires approximately 35 seconds per trial, which is faster than TuRBO(GP, NN) taking 40 seconds. With the first place in rank and first place in tavg, TuRBO(OP) is the Pareto frontier of speed and quality.

\begin{figure}
    \centering
    \includegraphics[width=1.0\linewidth]{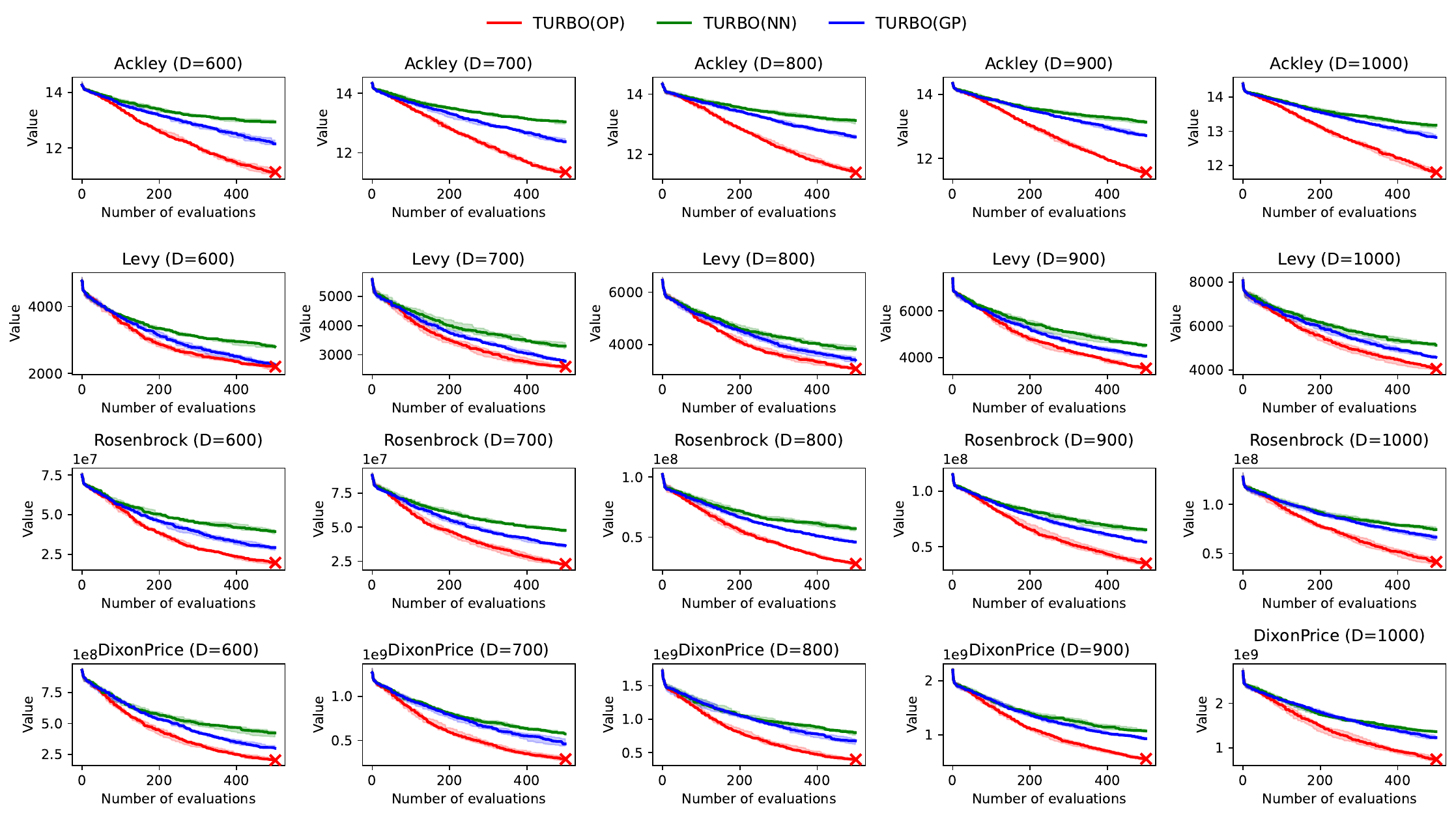}
    \caption{Optimization results on the synthetic benchmarks comparing our method against TuRBO algorithms.}
    \label{fig:res1}
\end{figure}

\begin{figure}
    \centering
    \includegraphics[width=1.0\linewidth]{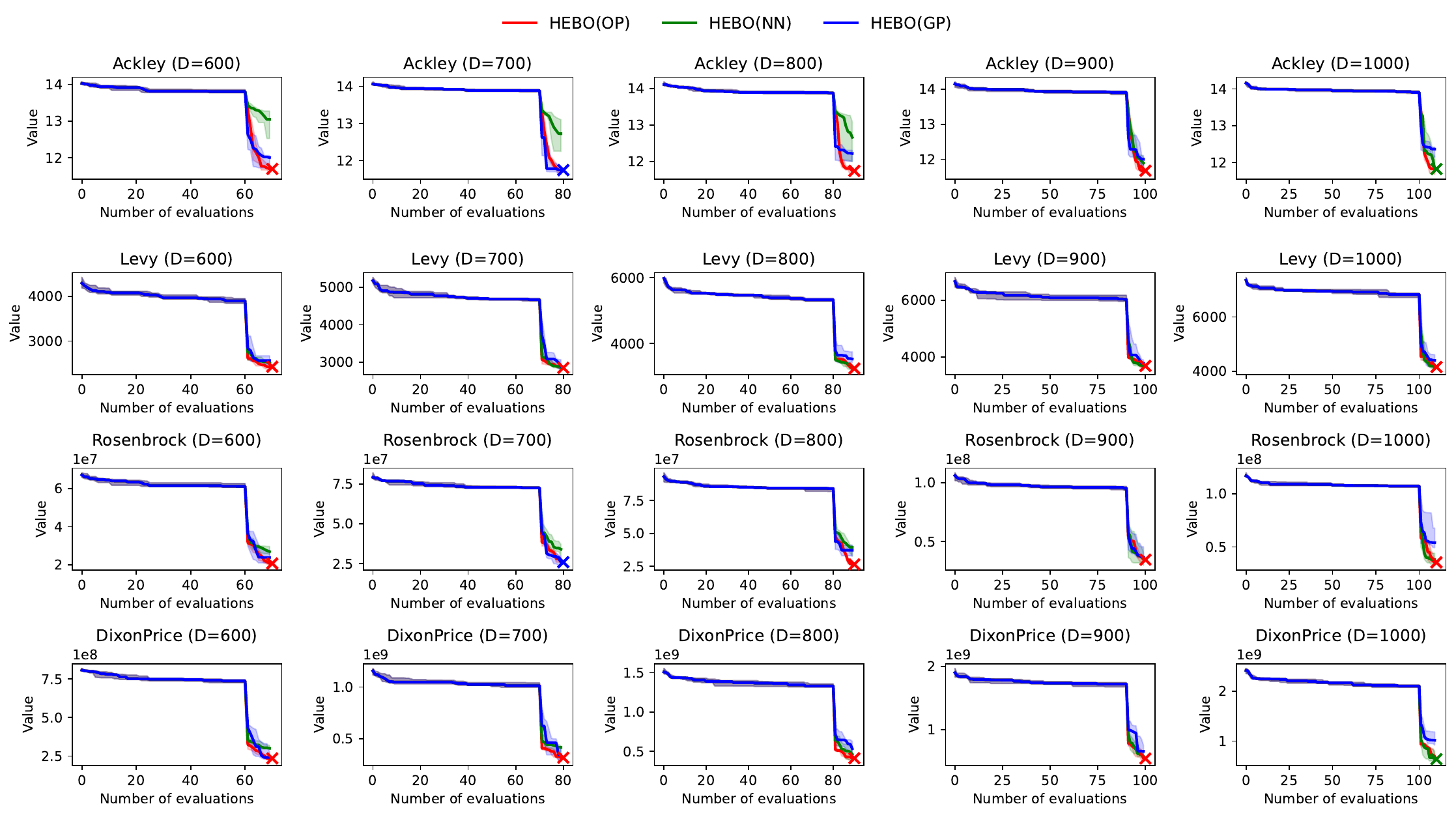}
    \caption{Optimization results on the synthetic benchmarks comparing our method against HEBO algorithms.}
    \label{fig:res2}
\end{figure}

\begin{figure}
    \centering
    \includegraphics[width=1.0\linewidth]{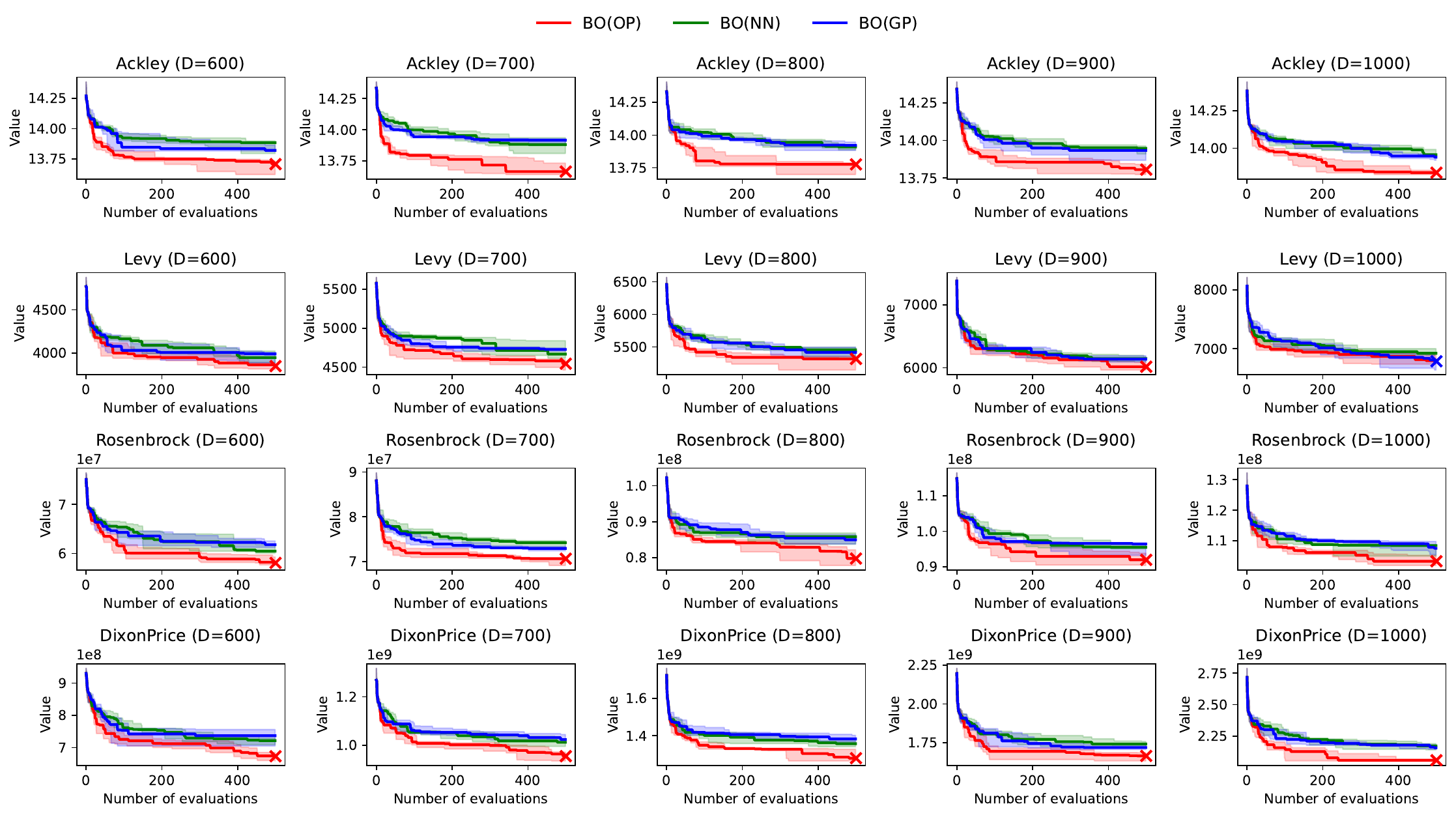}
    \caption{Optimization results on the synthetic benchmarks comparing our method against BO algorithms.}
    \label{fig:res3}
\end{figure}

\begin{figure}
    \centering
    \includegraphics[width=1.0\linewidth]{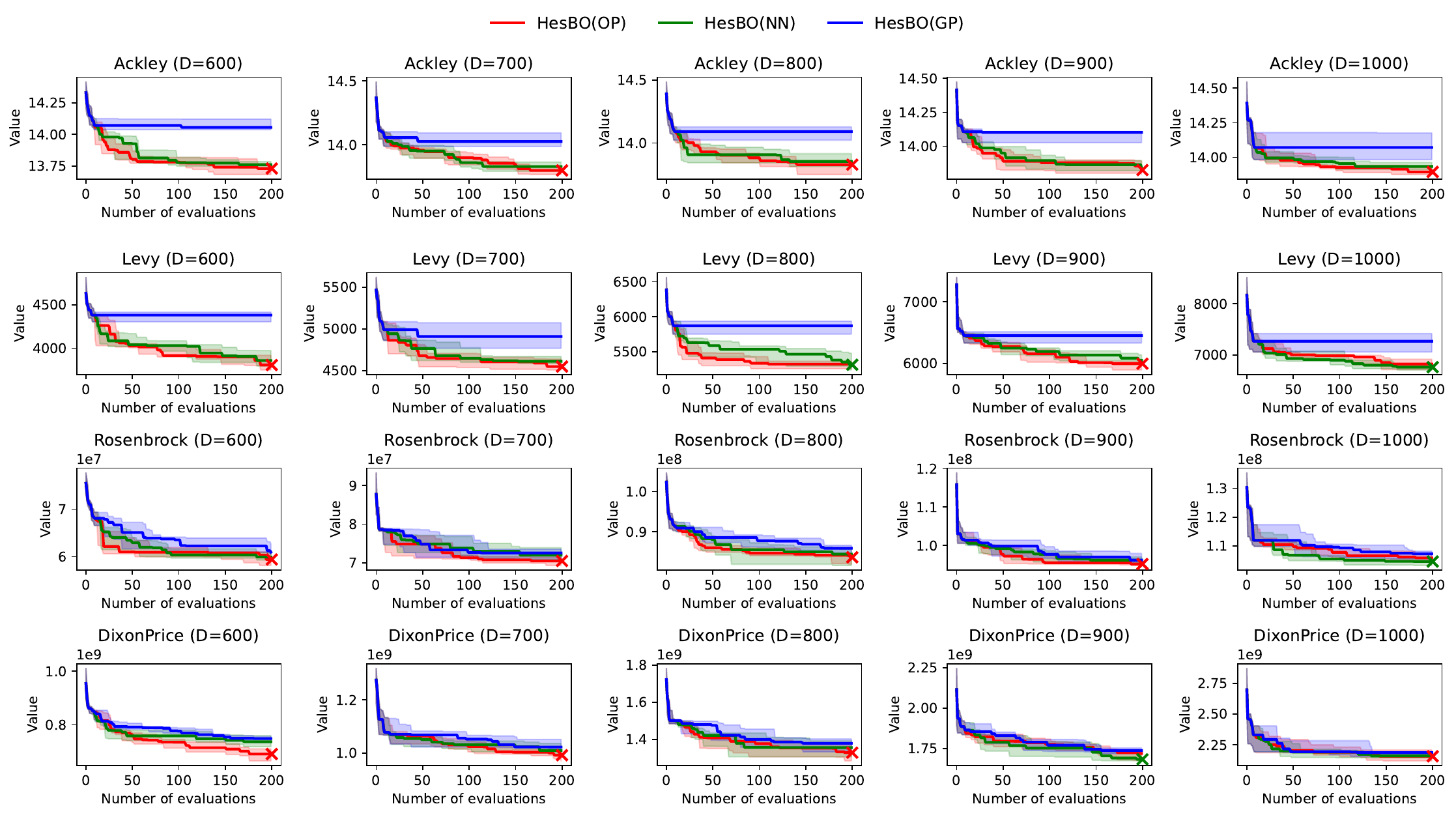}
    \caption{Optimization results on the synthetic benchmarks comparing our method against HesBO algorithms.}
    \label{fig:res4}
\end{figure}

\textbf{Scalable Synthetic Benchmarks (600 $\leq$ D $\leq$ 1000)} Overall, the performance of OP surpasses that of both NN and GP across all four baseline models. We present the comprehensive optimization results for all benchmark problems investigated in the main paper. Figures ~\ref{fig:res1} to ~\ref{fig:res4} provide the complete convergence curves for synthetic and real-world benchmarks, respectively. The results consistently demonstrate OP’s robust performance across different classes of high-dimensional problems, clearly indicating its advantage in both convergence speed and final solution quality compared to baseline methods. The convergence results for these scalable synthetic problems can be summarized in three categories. First, TuRBO(OP) has absolute domination over several problems (e.g., Ackley), outperforming all other surrogate models in all dimensions in Figure~\ref{fig:res1}. The second category, the most common observed ones, HEBO requires the fewest training epochs but the longest runtime, yet delivers excellent performance, making it particularly suitable for optimizing time-consuming functions in Figure~\ref{fig:res2}. Finally, HesBO(OP) struggles with very few problems (e.g., DixonPrice, D=600), where the current SOTA methods with the trust region dominate in Figure~\ref{fig:res4}.

\section{Conclusion}\label{sec_discussion}

In this paper, we introduce an OP surrogate model that significantly simplifies the modeling process by capturing relative order relationships instead of exact function values. The proposed OPBO relaxes the objective from finding the best to identifying a good-enough solution which achieves enhanced robustness. The interesting future directions include applying the proposed approach to various applications, handling black-box constraints by utilizing the OP approximation.

\bibliographystyle{unsrt}  
\bibliography{references}

\end{document}